%% file: draft.tex
\documentclass[10pt,twocolumn,letterpaper]{article}

\usepackage{cvpr}
\usepackage{times}
\usepackage{epsfig}
\usepackage{epstopdf}
\usepackage{float}
\usepackage{graphicx}
\usepackage{amsmath}
\usepackage{amsthm}
\usepackage{amssymb}
\usepackage[lined,boxed,commentsnumbered,ruled]{algorithm2e}
\usepackage{caption}
\usepackage{subcaption}

\SetCommentSty{mycommfont}
\input{tex/macros}

\usepackage[pagebackref=true,breaklinks=true,letterpaper=true,colorlinks,bookmarks=false]{hyperref}

\cvprfinalcopy 

\def\cvprPaperID{2052} 

\ifcvprfinal\fi

\begin{document}

\title{Pose and Shape Estimation with Discriminatively Learned Parts}

\author{Menglong Zhu\thanks{These authors contributed equally to this work} ~~~ Xiaowei Zhou\footnotemark[1]  ~~~  Kostas Daniilidis \\
Computer and Information Science, University of Pennsylvania \\
{\tt\small \{menglong,xiaowz,kostas\}@cis.upenn.edu}}

\maketitle

\begin{abstract}
\input{tex/abstract}
\end{abstract}

\maketitle

\section{Introduction}

\input{tex/introduction}

\section{Related Work}
\input{tex/related}


\input{tex/local}

\input{tex/global}

\section{Model Inference}

\input{tex/inference}

\section{Experiments}

\input{tex/experiment}

\section{Conclusion}

\input{tex/conclusion}

\bibliographystyle{ieee}
\small
\bibliography{bib/mybib,bib/menglongbib}

\end{document}

%% file: tex/macros.tex
\newcommand{\R}[1]{\mathbb{R}^{#1}}
\newcommand{\RR}[2]{\mathbb{R}^{#1 \times #2}}

\newcommand{\st}{\mbox{s.t.~}}

\newcommand{\refEq}[1]{(\ref{#1})}

\newcommand{\refSec}[1]{Section~\ref{#1}}

\def\bfr{{\mathbf{r}}}

\def\bft{{\mathbf{t}}}

\def\bfx{{\mathbf{x}}}

\def\bfone{{\mathbf{1}}}

\def\half{\frac{1}{2}~}

%% file: tex/abstract.tex
We introduce a new approach for estimating the 3D pose and the 3D shape of an object from a single image. Given a training set of view exemplars, we learn and select appearance-based discriminative parts which are mapped onto the 3D model from the training set through a facility location optimization. The training set of 3D models is summarized into a sparse set of shapes from which we can generalize by linear combination. Given a test picture, we detect hypotheses for each part. The main challenge is to select from these hypotheses and compute the 3D pose and shape coefficients at the same time. To achieve this, we optimize a function that minimizes simultaneously the geometric reprojection error as well as the appearance matching of the parts. We apply the alternating direction method of multipliers (ADMM) to minimize the resulting convex function. We evaluate our approach on the Fine Grained 3D Car dataset with superior performance in shape and pose errors. Our main and novel contribution is the simultaneous solution for part localization, 3D pose
and shape by maximizing both geometric and appearance compatibility.


%% file: tex/introduction.tex
Geometric features were the main representation in object recognition in the 20th century \cite{grimsonbook}.
Images of 3D objects were usually assumed to be segmented out and correspondence of well defined image features to projections of vertices or edges were established through voting for geometric consistency. Although such approaches were successful with geometric invariance they could not cope with the complexity of appearance of 3D objects in the real world which could only be learnt from exemplars. As soon as such 2D image exemplars became available in the Internet and through tedious annotation by the community, appearance based approaches exploded and the computer vision community is proud of the state of the art in detecting object categories via a bounding box or even segmenting them \cite{felzenszwalb2010object}. 
 Pose variation in 3D objects was converted into a 2D problem by clustering view exemplars into different classes \cite{hejrati2012analyzing,gu2010discriminative}. 
 Recently, researchers have realized that different views of the same 3D object can be 
married with existing 2D approaches like the Deformable Part Models by either extending the pictorial structures to three dimensions or rendering views of actual 3D CAD models \cite{pepik20123d2pm,pepik2012teaching,zia2013detailed,liebelt2008viewpoint}. 

We believe that three are the main challenges in the marriage of 2D appearance and 3D geometry: (1) how to learn a representation that efficiently predicts the appearance of geometric features given a pose and shape, (2) how to optimize for appearance and correspondence compatibility as well as 3D pose at the same time, without splitting the problem into subproblems of discrete poses, and (3) how to establish the 3D shape of an object when we want to avoid comparing serially with all possible 3D instances or when that instance has not been seen before. While 2D pictorial structures have been used in order to capture deformations for the sake of detection, we are genuinely interested in establishing the actual 3D shape of an object for the sake of fine grained classification or 3D interaction like grasping and manipulation.

In this paper, we propose a novel approach that marries the power of discriminative parts with an explicit 3D geometric representation with the goal to infer 3D pose as well as 3D shape of an object from a single image. We use the power of discriminative learning of parts to learn part descriptors in 2D training images enriched with the projection of a wired 3D model. Such parts are centered around projections of 3D landmarks which are given in abundance on the 3D model. To establish a compact representation we minimize the number of needed landmarks by solving a facility location linear program where the selection maximizes simultaneously discriminatively as well as the "serving" of the 3D landmarks that will be left out.

\begin{figure*}[htb]
\label{fig:highlight}
        \centering
        \includegraphics[width=\textwidth]{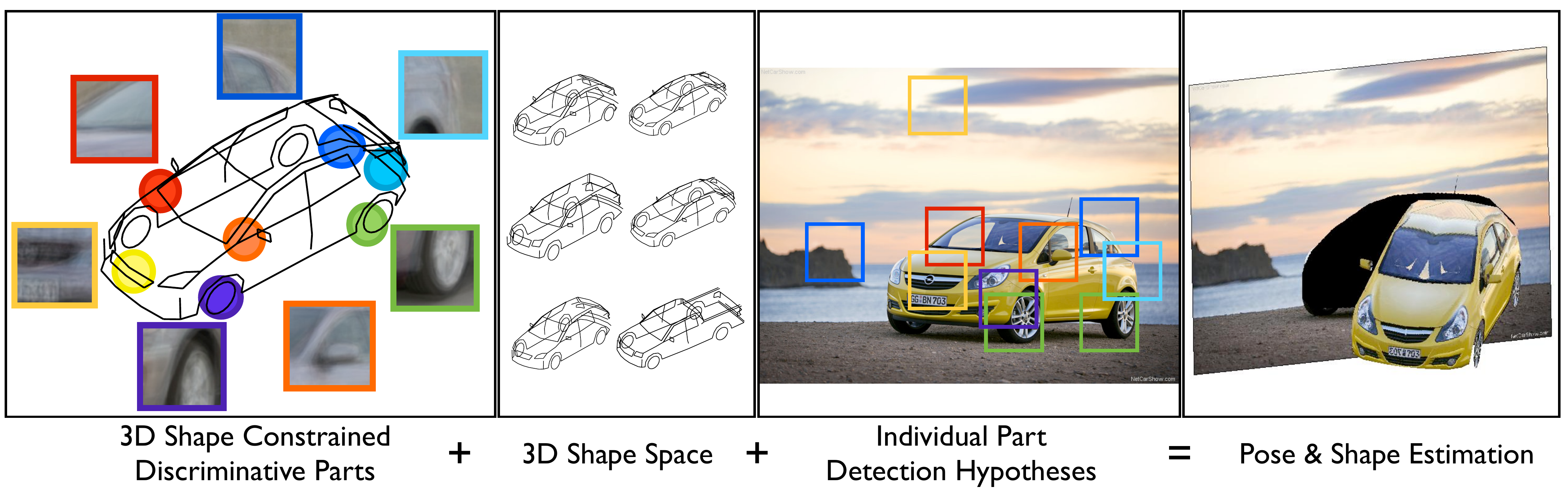}
             \caption{Illustrative summary of our approach: 3D Landmarks on a 3D model are associated with discriminatively learned part descriptors (left). Intra-class shape variation is captured with linear combinations of a sparse shape basis (2nd left). Learned part descriptors produce multiple maximum responses for each part in a testing image (3rd from left). The selection of the part hypotheses, 3D pose and 3D shape are simultaniously estimated and the result is illustrated through a popup (right).}\label{fig:highlight}
\end{figure*}

Given a learned part model for each landmark we detect top hypotheses for the location of each landmark in a testing image. The challenge is how to fit best these parts by maximizing the geometric consistency. This entails the selection among the hypotheses of each part and the pose/shape computation. Unlike other approaches which initialize pose by detecting DPM-based discretized poses \cite{zia2013detailed,Lin2014jointly},
we compute the selection as well as the 3D pose in one step using a mean-shape of the object category. We are able to achieve this by formulating as a convex optimization problem solvable by the alternating direction method of multipliers (ADMM). Subsequently, we apply two prunings of the hypotheses for each landmark projection. First, we prune by visibility induced by the estimated pose, and second we prune by proximity after solving for pose from the visible landmarks. One final application of the ADMM optimization solves for the pose as well as the shape. Joint pose and shape optimization is  achieved by joining the coefficients with respect to a sparse shape basis and the 3D rotation parameters into one matrix variable for each shape basis.

The main contributions of our approach are:
\begin{itemize}
\item
A compact learned representation of part descriptors corresponding to 3D landmarks.
\item
Resolving the 2D-3D marriage by simultaneously optimizing for appearance compatibility and geometric consistency. Unlike pairwise constraints in 2D pictorial structures or graphical models, here structure is formulated as global geometric consistency.
\item
Global geometric consistency does not mean rigidity, and unlike RANSAC based approached localizing 3D instances, we can deal with deformation by estimating the coefficients with respect to a shape basis.  
\item
Unlike approaches based on nonlinear minimization \cite{hejrati2012analyzing,Lin2014jointly,zia2013detailed}, we do not need initial estimates and we are not stuck on local minima. 
\end{itemize}

Our paper follows a classic organization, starting with the related work (Sec. 2), the learning of the representation in Sec. 3, the inference in Sec. 4, and results in Sec. 5. Figure \ref{fig:highlight} illustrates the outline of our approach.

%% file: tex/related.tex
Our model representation is inspired by recent advances in part-based modeling \cite{felzenszwalb2010object,Singh2012DiscPat,hariharan2012discriminative,kokkinos2011rapid}, which models the appearance of object classes with mid-sized discriminative parts.

The most popular approach to 3D object detection and viewpoint classification is to represent a 3D object by a collection of view-dependent 2D models separately trained on discretized views. Examples of this approach include \cite{schneiderman2000statistical,torralba2007sharing,felzenszwalb2010object,sun2009multi,gu2010discriminative,pepik2012teaching}. While these methods have shown superior detection performance, they provide relatively weak information about 3D geometry of objects. Some recent works directly used 3D models to encode the geometric relations among local parts and achieved continuous pose estimation \cite{yan20073d,savarese20073d,liebelt2008viewpoint,sun2010depth,glasner2011viewpoint,fidler20123d,xiang2012estimating,lim2013parsing,pepik20123d2pm,aubry2014seeing}. But they either used generic class models or instance-based models. Our approach differs in that we not only provide detailed shape representation but also consider intra-class variability.

The most related  category of methods is the one based on a shape-space model and tackling the recognition problem by aligning the shape model to image features. This approach originated from the active shape model (ASM) \cite{cootes1995active}, which was originally used for segmentation and tracking based on low-level image features. Cristinacce and Cootes \cite{cristinacce2006feature} proposed the constrained local models (CLM), which combined ASM with local appearance models for 2D feature localization in face images. Gu and Kanade \cite{gu20063d} presented a method to align 3D deformable models to 2D images for 3D face alignment. The similar methods were also proposed for 3D car modeling \cite{hejrati2012analyzing,zia2013detailed,Lin2014jointly} and human pose estimation \cite{ramakrishna2012reconstructing,zhou2014sptio}. Our method differs in that we use a data-driven approach for discriminative landmark selection and we solve landmark localization and shape reconstruction in a single convex framework, which enables the problem to be solved globally.


Our optimization approach is related to the previous work on using convex relaxation techniques for objet matching, e.g. \cite{maciel2003global,jiang2011linear,li2011optimal}. These methods focused on finding the point-to-point correspondence between an object template and a testing image in 2D, while our method considers 3D to 2D matching as well as shape variability.

%% file: tex/local.tex
\section{Shape Constrained Discriminative Parts}{\label{sec:discriminate}}

Our proposed method models both 2D appearance variation and 3D shape deformation of an object class. The 2D appearance is modeled as a collection of discriminatively trained parts. Each part is associated with a 3D landmark point on a deformable 3D shape. 


Unlike the previous works that manually define landmarks on the shape model, we propose an \textit{automatic} selection scheme: we first learn the appearance models for all points on the 3D model, evaluate their detection performance, and select a subset of them as our part models based on their detection performance in 2D and the spatial coverage in 3D.

\subsection{Learning Discriminative Parts}
One of the main challenges in object pose estimation rises from the fact that due to perspective transform and self occlusions, even the same 3D position of an object has very different 2D appearances in the image observed from different viewpoints. We tackle this problem by learning a mixture of discriminative part models for each point in the 3D model to capture the variety in appearance.

Given a training set $D$, each training image $I_i \in D$ is associated with the 3D points of the object shape $S \in \mathbb{R}^{3 \times p}$, their 2D projections $L_i \in \mathbb{R}^{2 \times p}$ annotated in the image, and their visibility $V_i \in \{0,1\}^p$.
For each visible 3D point $j \in \{1, \dots, p\}$, in training image $I_i$, we extract an $N \times N$ image patch centered at its 2D location $L_{ij}$ as positive example, assuming all the images are resized such that the object is approximately at the same scale. Negative examples are randomly extracted patches that do not have overlap with the object.

We bootstrap the learning of a discriminative mixture model for each part via clustering. Recent work \cite{hariharan2012discriminative,Singh2012DiscPat} has shown that whitened HOG (WHO) achieves better clustering results than HOG \cite{dalal2005histograms}. Denote $\phi(L_{ij})$ as the HOG feature of the positive image patch centered at $L_{ij}$ and $\overline{\phi}_\textbf{bg}$ as the mean of background HOG features. We compute the WHO feature as $\Sigma^{-1/2}(\phi(L_{ij}) - \overline{\phi}_\textbf{bg})$, where $\Sigma$ is the shared covariance matrix computed from all positive and negative features. Then we cluster the WHO features of each part $j$ into $m$ clusters using K-means as the initialization of the mixture model.

A linear classifier $W_{cj}$ is trained for each cluster $c$ of a part $j$. We apply linear discriminant analysis due to efficiency in training and limited loss in detection accuracy \cite{hariharan2012discriminative,girshick2013training},
\begin{align}\label{eq:lda}
W_{cj} = \Sigma^{-1}\left(\overline{\phi}(L_{ij};z_{ij} = c) - \overline{\phi}_\textbf{bg}\right),
\end{align}
where $z_{ij} \in \{1,\dots,m\}$ is the cluster assignment for each feature, and $\overline{\phi}(L_{ij};z_{ij} = c)$ is the mean feature over all $L_{ij}$ of cluster $c$. Let $\mathbf{x} = (x,y)$ be the position $(x,y)$ in the image. The response of part $j$ at a given location $\mathbf{x}$ is the max response over all its $c$ components: $score_j(\mathbf{x}) = \max_c \{W_{cj} \cdot \phi(\mathbf{x})\}$.

\begin{figure}
\centering
\includegraphics[width=.6\linewidth]{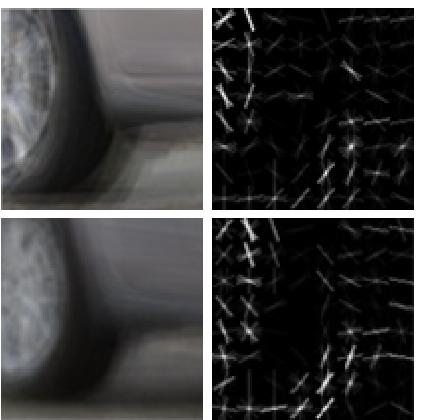}
\caption{The mean training patches and the positive weights of the learned filter for a (view) component of a part on the car wheel are shown. Filter learning is bootstrapped by clustering. The top and bottom rows correspond to the results before and after the latent update, respectively. The latent update procedure updates the center position and scale for each training patch by detecting the patches in the local region of the 2D landmark, and retrain the filters using better aligned training patches. The procedure results in more concentrated weights in learned filters.}\label{fig:latent}
\end{figure}

Due to intra-class variation and viewpoint differences, the appearance of training patches may not be perfectly aligned. Such misalignment results in inferior detector performance. We introduce a latent variable for each training patch, $r_{ij} \in \mathbb{R}^2$ to represent the relative center location to the annotated landmark location $L_{ij}$. To improve the classifier, we use the classifiers learned from (\ref{eq:lda}) to reposition the patch center in the neighborhood $\Delta(L_{ij})$ of $L_{ij}$ and update $z_{ij}$ and $r_{ij}$ as
\begin{align*}
\mathbf{x}^*_{ij} =&\; \text{argmax}_{\mathbf{x} \in \Delta(L_{ij})} score_j(\mathbf{x}),\\
z_{ij} =&\; \text{argmax}_{c} \{W_{cj} \cdot \phi(\mathbf{x}^*_{ij})\},\\
r_{ij} =&\; \mathbf{x}^*_{ij} - L_{ij}.
\end{align*}
The classifier weights are then retrained as
\begin{align}
\widehat{W}_{cj} = \Sigma^{-1}\left(\overline{\phi}(L_{ij}+r_{ij};z_{ij} = c) - \overline{\phi}_{\textbf{bg}}\right),
\end{align}
where $\overline{\phi}(L_{ij}+r_{ij};z_{ij} = c)$ is the mean feature over aligned patches of cluster $c$. Note that the latent update procedure is similar to that of DPM \cite{felzenszwalb2010object} with the difference that we do not apply generalized distance transform to filter responses but only consider maximum responses within a local region. The reason is that our model, as will be discussed in Section \ref{sec:global}, is constrained by the 3D shape space instead of learned 2D deformations. We want, thus, to obtain accurate part localization to estimate the object pose and shape.
Figure \ref{fig:latent} shows an example comparison of the mean image patch and the filter learned from clustering and after several iterations of latent update.
With the latent update step, the image patches of the training set become better aligned resulting in more concentrated weights in the learned filter. After the latent update, each patch mixture is retrained by hard negative mining and linear SVM as in \cite{felzenszwalb2010object} to boost detection accuracy.
A $2 \times 2$ covariance matrix $D_j$ is estimated for each landmark $j$ from latent variables $r_{ij}$, to model the uncertainty of the detected landmark position $\mathbf{x}^*_{ij}$  relative to the ground truth.

\subsection{Selecting Discriminative Landmarks}
Seeking a compact representation of the object, we try to select only a small subset of discriminative landmarks $S_D$ among all 3D landmarks $S$. We want the selected landmarks $S_D$ to be both associated with discriminative part models and have a good spatial coverage of the object shape model in 3D. The selection problem is formulated as a \textbf{facility location problem}, 
\begin{align}\label{eq:facility}
\min_{y_u, x_{uv}} ~~ & \sum_u z_{u} y_{u} + \lambda \sum_{uv} d_{uv} x_{uv}, \\
\st ~~ & \sum_v x_{uv} = 1,\notag\\
& x_{uv}  \le y_v, ~~~~~~~~~~~~~~~~ \forall u, v, \notag\\
& x_{uv}, y_u \in \{0,1\}, ~~~~~ \forall u, v, \notag
\end{align}
where the interpretations of each symbol are presented in Table \ref{table:facility}.
\begin{table}[h]
  \begin{tabular}{ll}
   \hline
    Symbol & Interpretation  \\
    \hline
    $z_u$ & cost of selecting landmark $u$ \\
    $y_u$ & binary landmark selection variable \\
    $d_{uv}$ & cost of landmark $v$ ``serving'' $u$ \\
    $x_{uv}$ & binary variable for landmark $v$ ``serving'' $u$ \\
    $\lambda$ & trade off between unary costs and binary costs\\
   \hline
\end{tabular}
  \caption{Notations interpretation in \refEq{eq:facility}}\label{table:facility}
\end{table}

\begin{figure}
\label{fig:selection}
\centering
\includegraphics[width=0.9\linewidth]{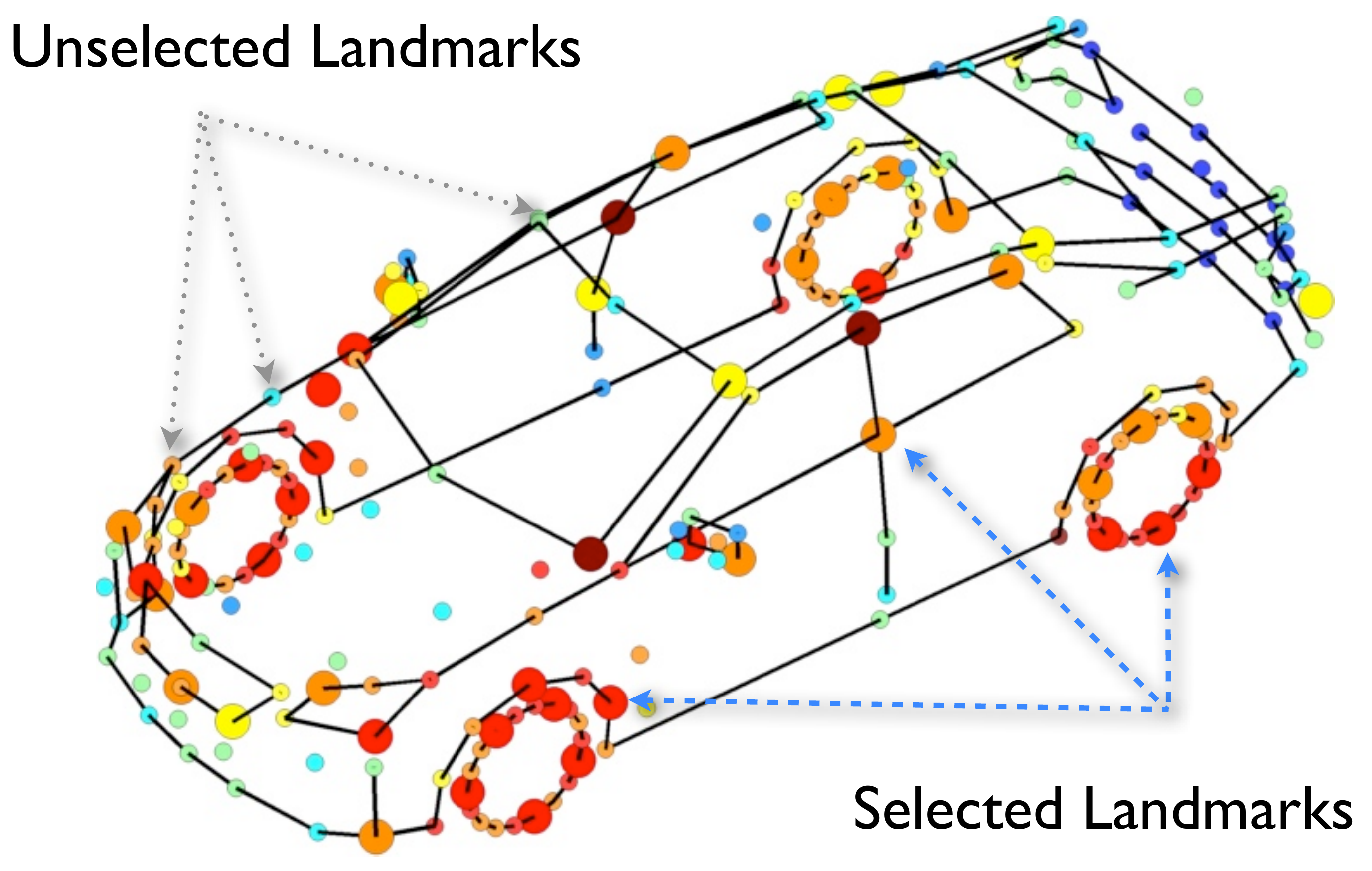}
\caption{Visualization of the landmark selection optimization result. All 256 landmark points of a car are shown in circle markers. The color of the markers represents the Average Precision(AP) of the landmark part detection on the training set, red means higher AP and blue means lower AP. The size of the landmark represents the selection result, the larger ones are selected via the MIP optimization and the smaller ones are not selected. The red landmarks are preferred since they have higher detection accuracy, but only a subset of red landmarks are selected because they are close in 3D.}
\label{fig:landmark-selection}
\end{figure}

The cost $z_u$ for a landmark $u$ should be lower if the associated part model is more discriminative. We model the discriminativeness by evaluating the Average Precision (AP) of detecting each landmark in the training set. For any landmark $u$, we perform detection with the learned part model in the training set $S$ to generate a list of location hypotheses $H_u$. A hypothesis $h \in H_u$ is considered as true positive if the ground truth location $L_{iu}$ is within a small radius $\delta$. Let the computed AP for a part $u$ be $AP_u$, we set $z_u = 1 - AP_u$. The cost of ``serving'' (or suppressing) other landmarks are set to be the euclidean distance between landmarks in 3D, i.e., $d_{uv} = ||S_u - S_v||_2$. The value of $\lambda$ is set to 1 in our experiments.
The minimization problem \ref{eq:facility} is a Mixed Integer Programming (MIP) problem, which is known to be NP-hard. But a good approximation solution can be obtained by relaxing the integer constrains to be $x_{uv} \in [0,1], ~ y_{u} \in [0,1]$, solving the relaxed Linear Programming problem, and thresholding the solution.
Figure \ref{fig:landmark-selection} visualizes an example result of MIP optimization for landmark selection.

%% file: tex/global.tex
\subsection{3D Shape Model}\label{sec:global}

We start our description by explaining how we would estimate the pose and shape of an object if 2D part - 3D landmark correspondences were known. We represent a 3D object model as a linear combination of a few basis shapes to constrain the shape variability. This assumption has been widely used in various shape-related problems such as object segmentation \cite{cootes1995active}, single image-based shape recovery \cite{gu20063d} and nonrigid structure from motion \cite{bregler2000recovering}. We use a weak-perspective model, which is a good approximation when the depth of the object is smaller than the distance from the camera. With these two assumptions, the 2D part locations $P\in\RR{2}{p}$ can be described by
\begin{align}\label{eq:shape1}
P = R\sum_{i=1}^{k} c_iB_i + \bft\bfone^T,
\end{align}
where $B_i\in\RR{3}{p}$ denotes the $i$-th basis shape, $R\in\RR{2}{3}$ represents the first two rows of a rotation matrix, and $\bft\in\R{2}$ is the translation vector. In model inference, we try to minimize the geometric reprojection error to find the optimal parameters.

However, the model in \refEq{eq:shape1} is bilinear in $R$ and $c_i$s yielding  a nonconvex problem. In order to have a linear representation, we use the shape model proposed in \cite{zhou20143d}, which assumes that there is a rotation for each basis shape. The 3D shape model is $S=\sum_{i=1}^{k}c_iR_iB_i$, and the 2D part locations are given by
\begin{align}\label{eq:shape2}
P = \sum_{i=1}^{k} T_iB_i + \bft\bfone^T,
\end{align}
where $T_i\in\RR{2}{3}$ corresponds to the first two rows of $R_i$ multiplied by $c_i$. In order to enforce $T_i$ to be orthogonal, the spectral norms of $T_i$s are minimized during model inference. The spectral norm is the largest singular value of a matrix, and minimizing it enforces the two singular values to be equal, which yields an orthogonal matrix \cite{zhou20143d}. After $T_i$s are estimated, $c_i$s and $R_i$s are derived from $T_i$s and the shape is reconstructed by $S=\sum_{i=1}^{k}c_iR_iB_i$. Note that the reconstructed shape is in the camera frame, and we compute a single rotation matrix $R$ by aligning the reconstructed shape to the canonical pose.

%% file: tex/inference.tex
Finally, we obtain global geometry-constrained local-part models, in which the unknowns are the 2D part locations as well as the 3D pose and shape. In model inference, we maximize the detector responses over the part locations while minimizing the geometric reprojection error.

\subsection{Objective Function}\label{sec:obj}

We try to locate a part by finding its correspondence in a set of hypotheses given by the trained detector. The cost without geometric constraints is
\begin{align}
f_{score}(\bfx_1,\cdots,\bfx_p) = - \sum_{j=1}^{p} \bfr_j^T\bfx_j,
\end{align}
where $\bfx_j\in\{0,1\}^l$ is the selection vector and $\bfr_j\in\R{l}$ is the vector of the detection scores for all hypotheses for the $j$-th part.

Geometric consistency is imposed by minimizing the following reprojection error:
\begin{align}
&f_{geom}(\bfx_1,\cdots,\bfx_p,T_1,\cdots,T_k,\bft) = \nonumber\\
&\half \sum_{j=1}^{p} \left\| D_j^{-\frac{1}{2}}\left(L_j^T\bfx_j - \left[\sum_{i=1}^{k} T_iB_i\right]_j -\bft \right) \right\|^2,
\end{align}
where we concatenate the 2D locations of hypotheses for part $j$ in $L_j\in\RR{l}{2}$ and denote the covariance estimated in training as $D_j$.

As introduced in \refSec{sec:global}, we add  the following regularizer to enforce the orthogonality of $T_i$:
\begin{align}
f_{reg}(T_1,\cdots,T_k) = \sum_{i=1}^{k}\|T_i\|_2,
\end{align}
where we use $\|T_i\|_2$ to represent the spectral norm of $T_i$, i.e., the largest singular value.

To simplify the computation, we relax the binary constraint on $\bfx_i$ and allow it to be a soft-assignment vector $\bfx_i \in \mathcal{A}$, where $\mathcal{A}=\{\bfx\in[0.1]^l|~ \sum_{i=1}^{l}x_i=1.\}$.

Finally, the objective function reads
\begin{align}\label{eq:finalcost}
    \min_{\overline{X},\overline{T},\bft} ~& f_{geom}(\overline{X},\overline{T},\bft) + \lambda_1 f_{score}(\overline{X}) + \lambda_2 f_{reg}(\overline{T}), \\
    \st ~~~~ & \bfx_j\in\mathcal{A}, ~ \forall j=1:p, \nonumber
\end{align}
where $\overline{X}$ and $\overline{T}$ represent the unions of $\bfx_1,\cdots,\bfx_p$ and $R_1,\cdots,R_k$, respectively.
After solving \refEq{eq:finalcost}, we recover the 3D shape $S$ and pose $\theta=(R,\bft)$ from $T_i$s, as introduced in \refSec{sec:global}.

\subsection{Optimization}


The problem in \refEq{eq:finalcost} is convex since $f_{score}$ is a linear term, $f_{geom}$ is the sum of squares of linear terms, and $f_{reg}$ is the sum of norms of unknown variables. We use the alternating direction method of multipliers (ADMM) \cite{boyd2010distributed} to solve the convex problem in \refEq{eq:finalcost}. Since $f_{reg}$ is nondifferentiable, which is not straightforward to optimize, we introduce an auxiliary variable $Z$ and reformulate the problem as follows:
\begin{align}
    \min_{\overline{X},\overline{T},\bft,\overline{Z}} ~& f_{geom}(\overline{X},\overline{T},\bft) + \lambda_1 f_{score}(\overline{X}) + \lambda_2 f_{reg}(\overline{Z}), \\
    \st ~~~~ & \overline{T} = \overline{Z}, \nonumber \\
    & \bfx_j\in\mathcal{A}, ~ \forall j=1:p. \nonumber
\end{align}

The corresponding augmented Lagrangian is:
\begin{align}\label{eq:lagragian}
\mathcal{L} &= f_{geom}(\overline{X},\overline{T},\bft) + \lambda_1 f_{score}(\overline{X}) + \lambda_2 f_{reg}(\overline{Z}) \nonumber \\
&+ \left<Y,\overline{T}-\overline{Z}\right> + \frac{\rho}{2}\|\overline{T}-\overline{Z}\|_F^2.
\end{align}
The ADMM algorithm iteratively updates variables by the following steps to find the stationary point of \refEq{eq:lagragian}:
\begin{align}
\bft &\leftarrow \arg\min_{\bft} \mathcal{L}, \label{eq:step-t}\\
\overline{X} &\leftarrow \arg\min_{\overline{X}} \mathcal{L}, \label{eq:step-X}\\
\overline{T} &\leftarrow \arg\min_{\overline{T}} \mathcal{L}, \label{eq:step-T}\\
\overline{Z} &\leftarrow \arg\min_{\overline{Z}} \mathcal{L}, \label{eq:step-Z}\\
Y &\leftarrow \rho(\overline{T}-\overline{Z}).
\end{align}
It can be shown that \refEq{eq:step-t}, \refEq{eq:step-X} and \refEq{eq:step-T} are all quadratical programming problems, which have closed-form solution or can be solved efficiently using existing convex solvers. \refEq{eq:step-Z} is a spectral-norm regularized proximal problem, which also admits a closed-form solution \cite{zhou20143d}.

\subsection{Visibility Estimation}\label{sec:visibility}

In model inference, only visible landmarks should be considered.  To estimate the unknown visibility, we adopt the following strategy. We first assume that all landmarks are visible and solve our model in \refEq{eq:finalcost} to obtain a rough estimate of the viewpoint. Since the landmark visibility of a car only depends on the aspect graph, the roughly estimated viewpoint can give us a good estimate of the landmark visibility. We observed that our model could reliably estimate the coarse view by assuming the full visibility, which might be attributed to the global optimization. After obtaining the visibility, we solve our model again by only considering the visible landmarks. The full shape can be reconstructed by the linear combination of full meshes of basis shapes after the coefficients are estimated.

\subsection{Successive Refinement}\label{sec:prune}

The relaxation of binary selection vectors $\bfx_j$s in \refEq{eq:finalcost} may yield inaccurate localization, since it allows the landmark to be located inside the convex hull of the hypotheses. To improve the precision, we apply the following scheme: we solve our model in \refEq{eq:finalcost} repeatedly, and in each iteration we define a trust region based on the previous result for each landmark and merely keep the hypotheses inside the trust region as the input to fit the model again. We use three iterations. We can start from a large trust region to achieve global fitting and gradually decrease the trust region size in each iteration to reject outliers and improve localization. This successive refinement scheme has been widely-used for feature matching \cite{li2011optimal,jiang2011linear}.

We summarize the inference process in Algorithm \ref{alg:inference}.
\begin{algorithm}\small
\textbf{Input:} 2D hypotheses for parts $\{\mathcal{H}_i=(L_i,\bfr_i)|~i=1:m\}$ \\
3D basis shapes $\{B_i|~i=1:k\}$ and mean shape $B_0$\\
\textbf{Output:} Estimated pose $\theta=(R,\bft)$ and shape $S$\\
\tcc{$optimize(\{\mathcal{H}_i\},\{B_i\})$ means solving \refEq{eq:finalcost} with hypotheses $\{\mathcal{H}_i\}$ and basis shapes $\{B_i\}$ to recover $\theta$ and $S$ (\refSec{sec:obj})}
\tcc{Estimate visibility (\refSec{sec:visibility})}
$\theta,S ~ \leftarrow ~ optimize(\{\mathcal{\mathcal{H}}_i|~i=1:m\},B_0)$\;
$V ~ \leftarrow ~ visibility(\theta)$\;
\tcc{Prune hypotheses (\refSec{sec:prune})}
$\theta,S ~ \leftarrow ~ optimize(\{\mathcal{H}_i~i\in V\},B_0)$\;
$\tilde{\mathcal{H}}_i ~ \leftarrow ~  pruning(\mathcal{H}_i,S_i),\forall i\in V$\;
\tcc{Refine pose \& shape with shape space}
Update $\theta,S ~ \leftarrow ~ optimize(\{\tilde{\mathcal{H}}_i|~i\in V\},\{B_i,|~i=1:k\})$\;
\caption{Outline of the inference process}
\label{alg:inference}
\end{algorithm}

%% file: tex/experiment.tex
In this section, we evaluate our method (PSP) in terms of both pose and shape estimation accuracy. The experiments are carried out on the Fine Grained 3D Car dataset (FG3DCar) \cite{Lin2014jointly}, since it is the only dataset with both landmark projection in the image and pose annotation for 3D objects. The dataset consists of 300 images with 30 different car models of 6 car types under different viewing angles. Each car instance is associated a shape model of 256 3D landmark points and their projected 2D locations annotated in the image as well as 3D pose annotation. We perform the following evaluations: First, we compare the accuracy of pose and shape estimation to the iterative model fitting method of \cite{Lin2014jointly} (FG3D) in terms of 2D landmark projection error. Second, we compare the coarse viewpoint estimation error to viewpoint-DPM (V-DPM) \cite{gu2010discriminative,xiang_wacv14}. In addition, since our viewpoint estimation is continuous, we also show the angular errors comparing to the groundtruth annotation. Through out the experiments, we follow the same training-testing split as \cite{Lin2014jointly}, half of the images are used for training and half for testing. We increase the training set size by left-right flip the training images and using symmetry to flip the landmark visibility labels.

During training we learn a mixture of discriminative part models of three components for each of 256 landmark points as described in Section \ref{sec:discriminate}. The Average Precision (AP) of the landmark detection is evaluated on the training set. We count a detection as true positive only if the detected landmark location is within 20 pixels distance to the annotated location, otherwise it is counted as a false positive. We optimize the landmark selection with unary cost as 1 - AP of each landmark and pairwise cost as the average pairwise 3D distance over all the 3D models in the training set. 52 out of 256 landmark points are selected resulting from the MIP optimization. To build the shape models, we learned a dictionary consisting of 10 basis shapes from the 3D models provided in the FG3DCar dataset. We use $\lambda_1 = 1000$ and $\lambda_2 = 50$ in (\ref{eq:finalcost}) during inference.

Note that, unlike FG3D, our method does not need an external object detector to initialize either the location and scale in the image or coarse landmark locations. We perform pose and shape estimation on the original image with background clutter.

\begin{table}
    \centering
\begin{tabular}{|l|c|c|}
\hline
Method & meanAPD (SL) & meanAPD \\
\hline
PSP ~~ Mean shape  & 16.5 & 20.6\\
PSP ~~ Class mean & 15.4 & 18.9 \\
PSP ~~ Shape space  & \textbf{14.6} & \textbf{17.7} \\
\hline
FG3D Class mean & -  & 18.1 \\
FG3D Shape space & - & 20.3  \\
\hline
\end{tabular}
\caption{Model fitting error of PSP versus FG3D in terms of mean APD in pixels evaluated on 52 selected discriminative landmarks (SL) and 64 landmarks provided in the dataset.}
\label{table:fitting}
\end{table}

\begin{figure}
    \includegraphics[width=\linewidth]{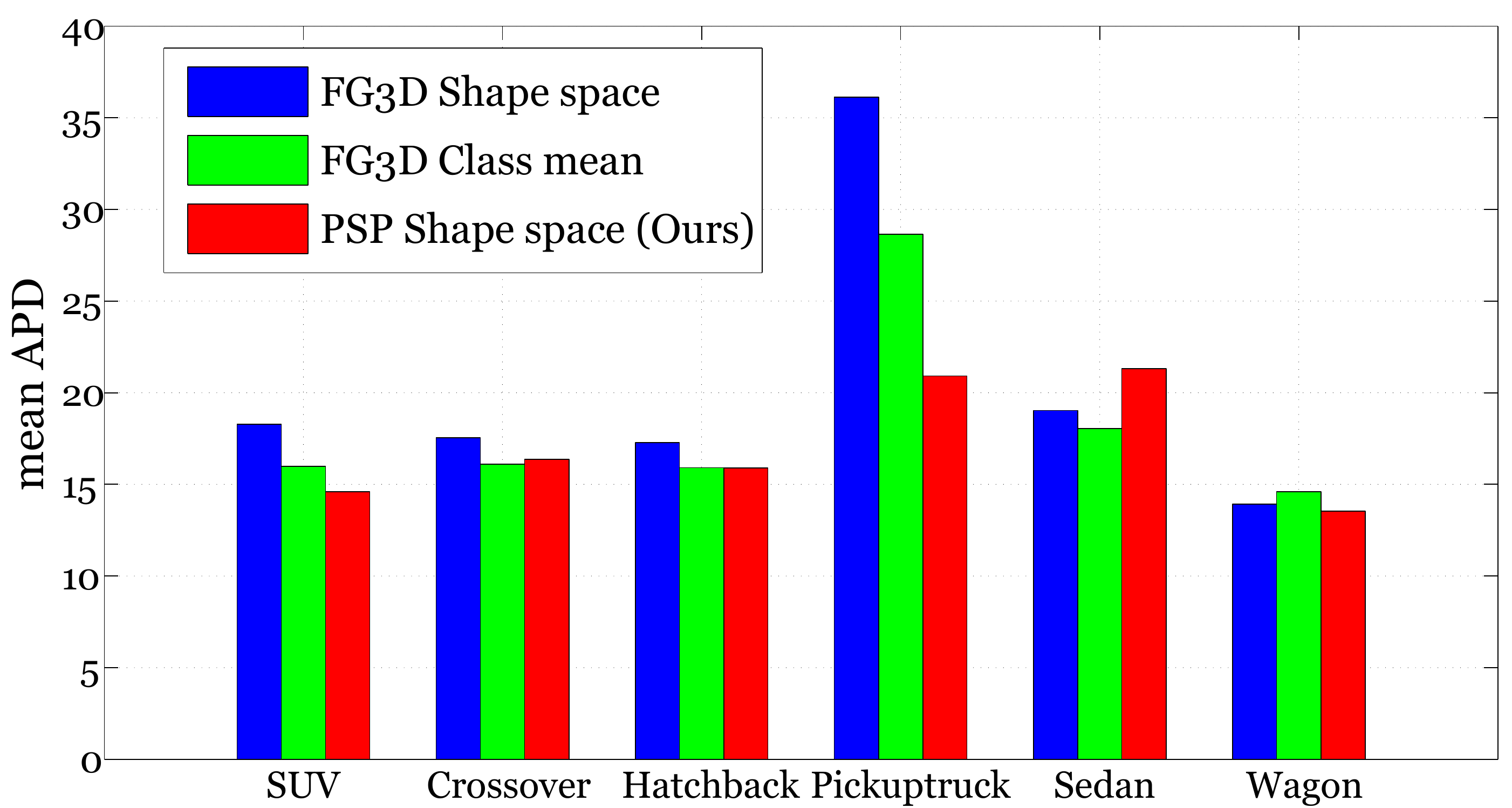}
    \caption{Car type specific meanAPD of PSP versus FG3D with mean prior and class prior. Comparing to FG3D method, our method achieves lower meanAPD on most car types. For the type of pickup truck, our method significantly outperforms FG3D.}
    \label{fig:error-bar}
\end{figure}

\begin{table}
    \centering
\begin{tabular}{|l|c|c|}
\hline
 & \multicolumn{2}{|c|}{Accuracy}  \\
\hline
 Method & $40^{\circ}$ per view & $20^{\circ}$ per view \\
\hline
V-DPM & 82.7\% & 71.3\% \\
PSP   & 89.3\% & 84.7\% \\
\hline
\end{tabular}
\caption{Coarse viewpoint estimation accuracy versus V-DPM evaluated on FG3DCar dataset. Accuracies are compared with two discretization schemes, 20 degrees per coarse viewpoint and 40 degrees per coarse viewpoint.}
\label{table:vp}
\end{table}

\begin{figure}
  \centering
  \includegraphics[width=\linewidth]{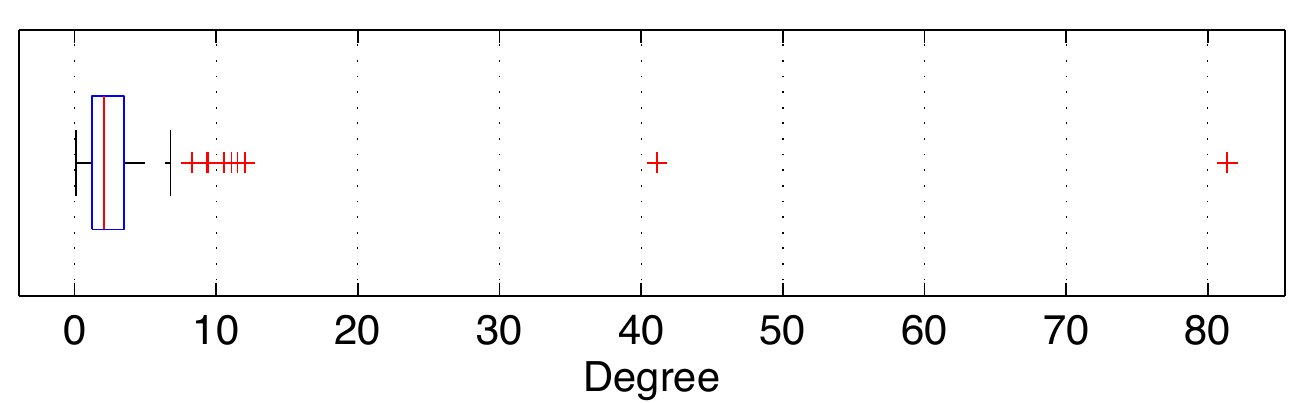}\\
  \caption{Continuous viewpoint (azimuth) error comparing to the groundtruth on all 150 test images in FG3DCar dataset. The mean error is 3.4 in degrees.}\label{fig:err-pose}
\end{figure}

\paragraph{3D Pose and Shape Estimation}
Pose and shape estimation accuracy is evaluated in terms of meanAPD which is the average landmark projection error in pixels over the landmarks and the test instances. In the following experiments, we investigate the effect of using different 3D shapes on the model fitting error. We compare three setups with different basis shapes: mean shape, mean class shape and shape space (10 basis shapes). The middle column of Table \ref{table:fitting} shows the fitting error on selected discriminative landmarks. The fitting error decreases when we use shape space instead of both mean shape and class mean, which validates the use of shape space to express intra-class shape variation.

Since the selected discriminative landmarks are not identical to the landmarks provided in the FG3DCar dataset, we also compare the meanAPD on the landmarks provided in the dataset. Our method outperforms FG3D using the shape space without knowing the class type.
Note that, their detectors are trained on the manually selected 64 landmarks provided in the dataset while our detectors are trained on the 52 automatic selected discriminative landmarks.

Although our objective is to optimize the projection error on the discriminative landmarks, the fitting error on the dataset provided landmarks is also minimized. This shows the effectiveness of the landmark selection process. The error is reported on the same scale as FG3D. Figure \ref{fig:error-bar} shows the per class 3D model fitting error. Our method outperforms FG3D on most class types with particular success on the pickup trucks.

\paragraph{Viewpoint Estimation}
We compare PSP to V-DPM in discrete viewpoint estimation accuracy. For V-DPM we train two sets of baseline V-DPM with coarse viewpoints (azimuth) of every 20 degrees and every 40 degrees for each view. Each component of V-DPM corresponding to a viewpoint label. During inference, the viewpoint of the test car instance is predicted as the training viewpoint of the max scoring component. For PSP, the estimated continuous viewpoint is discretized the same way as V-DPM. Table \ref{table:vp} shows the comparison of the two methods. In both two cases, PSP outperform V-DPM. We further analyze the estimation error of PSP by looking at continuous viewpoint estimation error and show that the majority error is introduced by discretization. We compare our estimation to ground-truth viewpoint (azimuth)  and report the absolute angular value in Figure \ref{fig:err-pose}. The mean error over the whole test set is only 3.4 in degree. 

\begin{figure*}
        \centering
         \begin{subfigure}[b]{0.25\textwidth}
                \includegraphics[width=\textwidth]{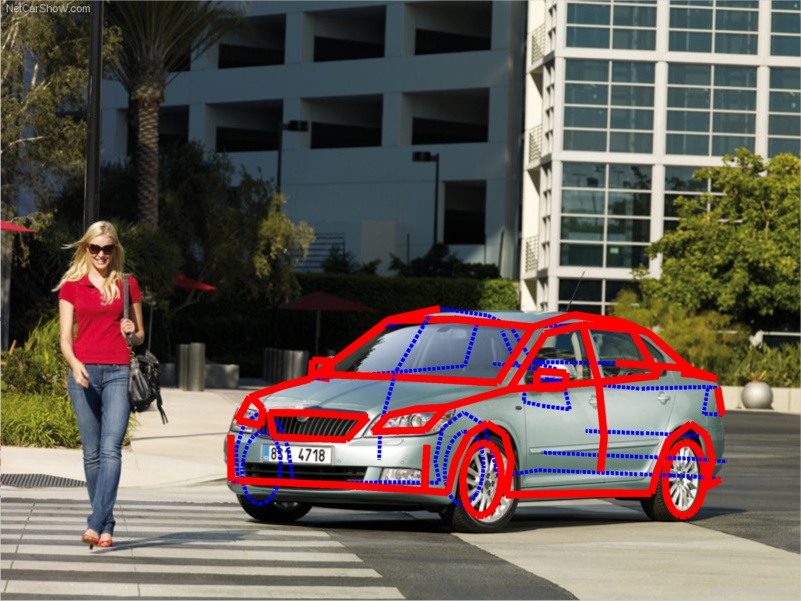}
        \end{subfigure}%
        ~ 
        \begin{subfigure}[b]{0.25\textwidth}
                \includegraphics[width=\textwidth]{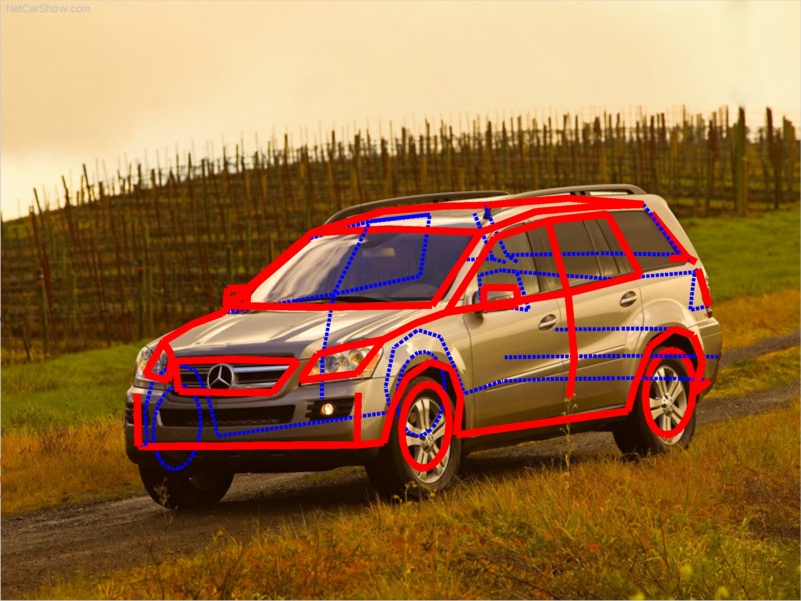}
        \end{subfigure}%
        ~ 
        \begin{subfigure}[b]{0.25\textwidth}
                \includegraphics[width=\textwidth]{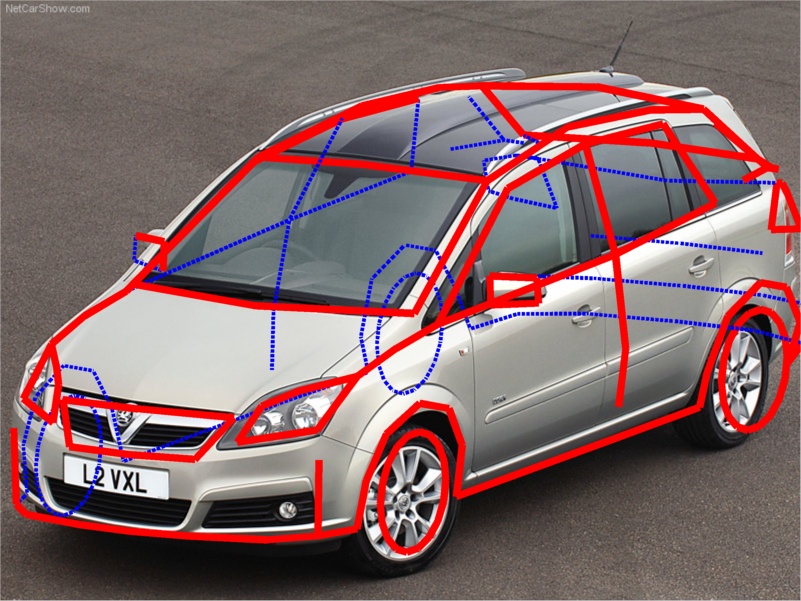}
        \end{subfigure}%
        ~
        \begin{subfigure}[b]{0.25\textwidth}
                \includegraphics[width=\textwidth]{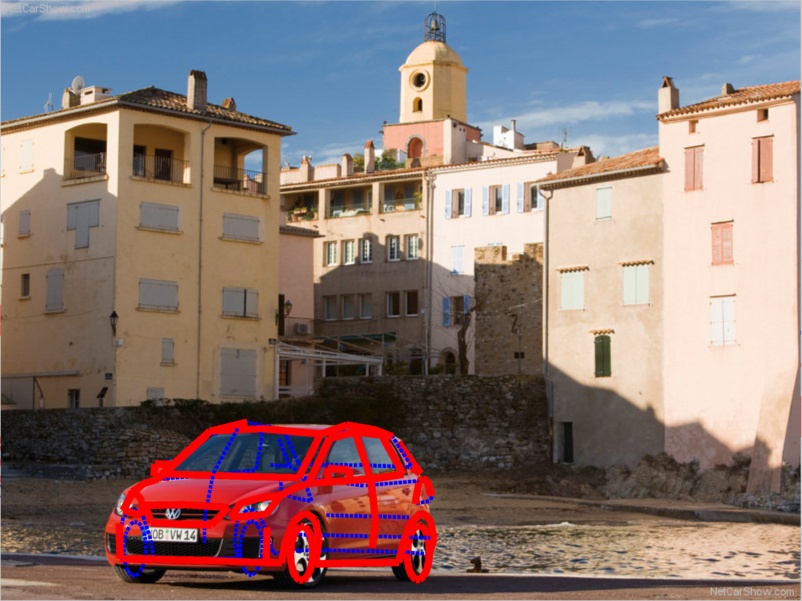}
        \end{subfigure}\\
        \begin{subfigure}[b]{0.25\textwidth}
                \includegraphics[width=\textwidth]{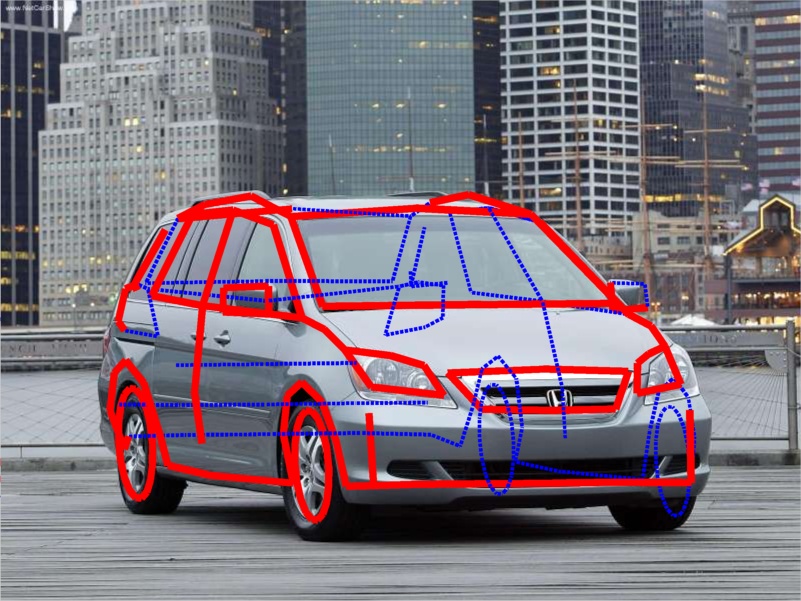}
        \end{subfigure}%
        ~ 
        \begin{subfigure}[b]{0.25\textwidth}
                \includegraphics[width=\textwidth]{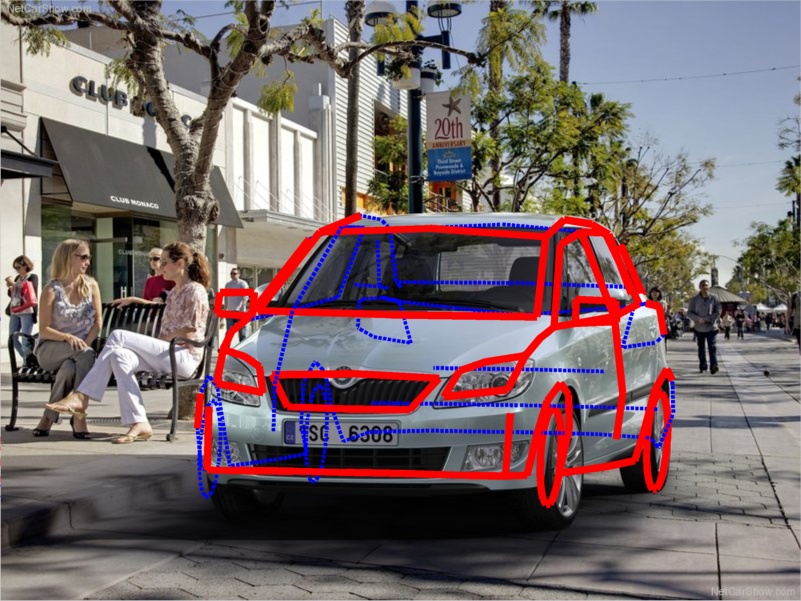}
        \end{subfigure}%
        ~ 
        \begin{subfigure}[b]{0.25\textwidth}
                \includegraphics[width=\textwidth]{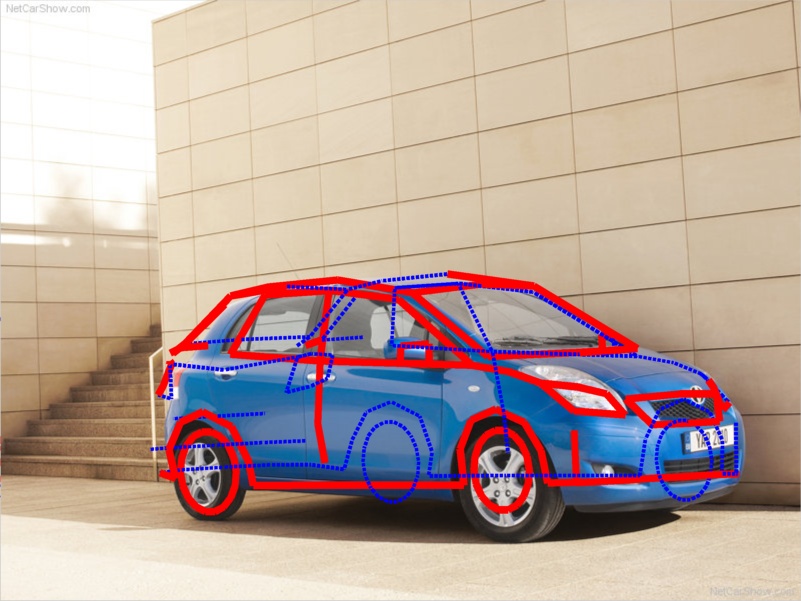}
        \end{subfigure}%
        ~
        \begin{subfigure}[b]{0.25\textwidth}
                \includegraphics[width=\textwidth]{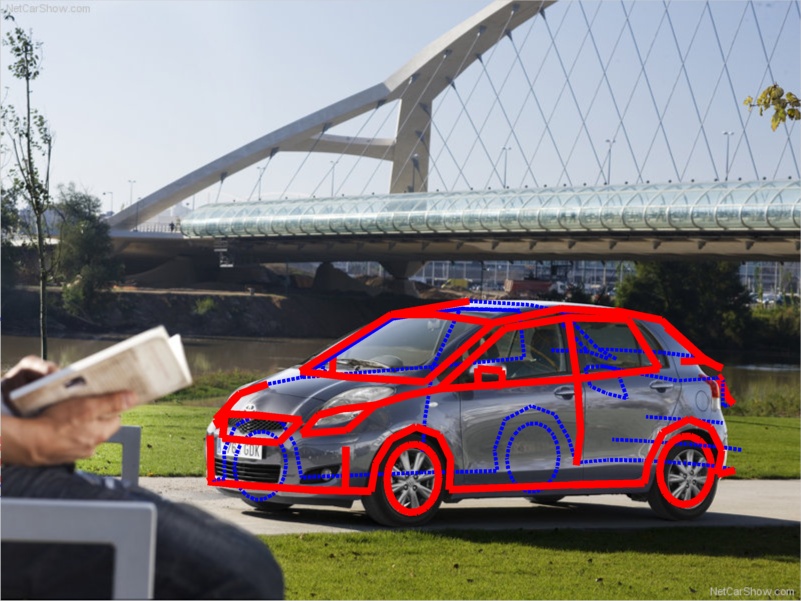}
        \end{subfigure}\\
        \begin{subfigure}[b]{0.25\textwidth}
                \includegraphics[width=\textwidth]{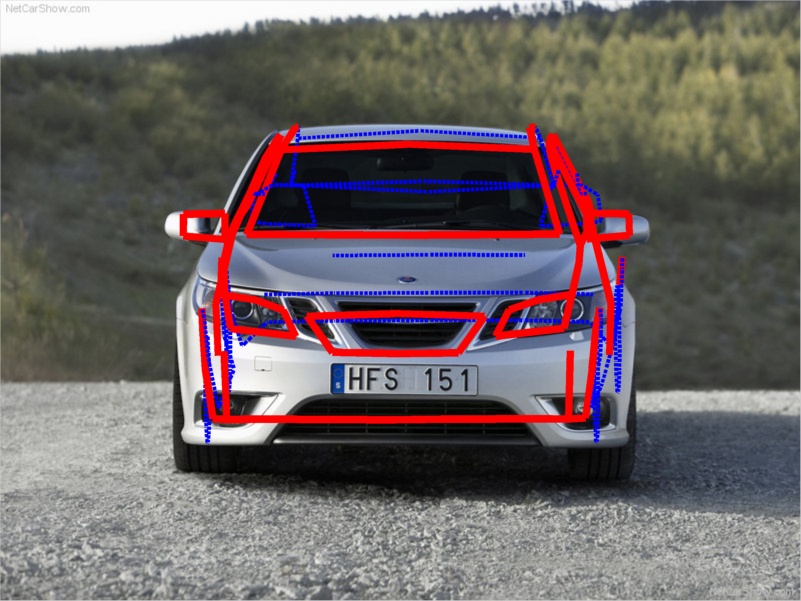}
        \end{subfigure}%
        ~ 
        \begin{subfigure}[b]{0.25\textwidth}
                \includegraphics[width=\textwidth]{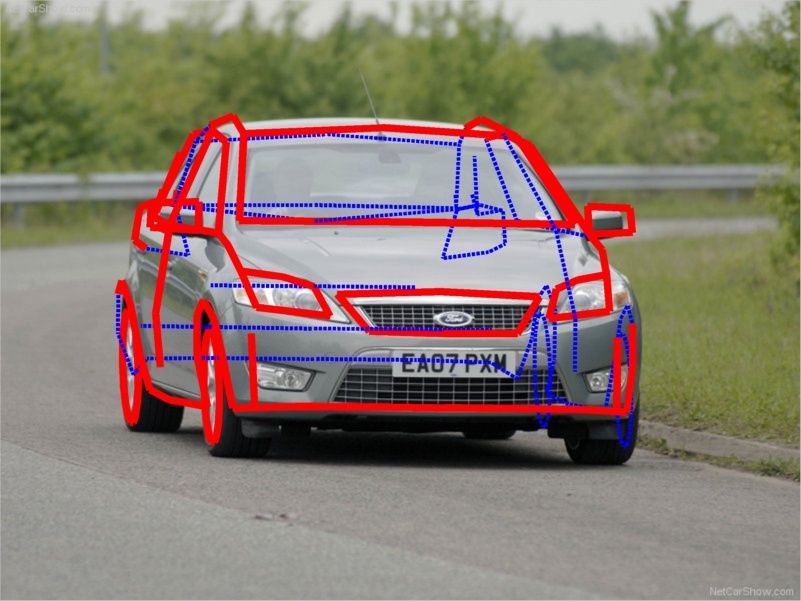}
        \end{subfigure}%
        ~ 
        \begin{subfigure}[b]{0.25\textwidth}
                \includegraphics[width=\textwidth]{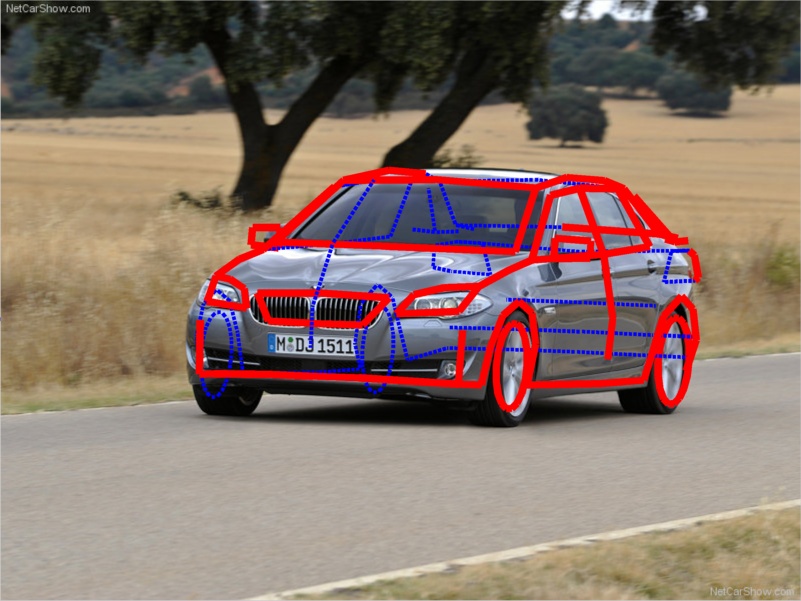}
        \end{subfigure}%
        ~
        \begin{subfigure}[b]{0.25\textwidth}
                \includegraphics[width=\textwidth]{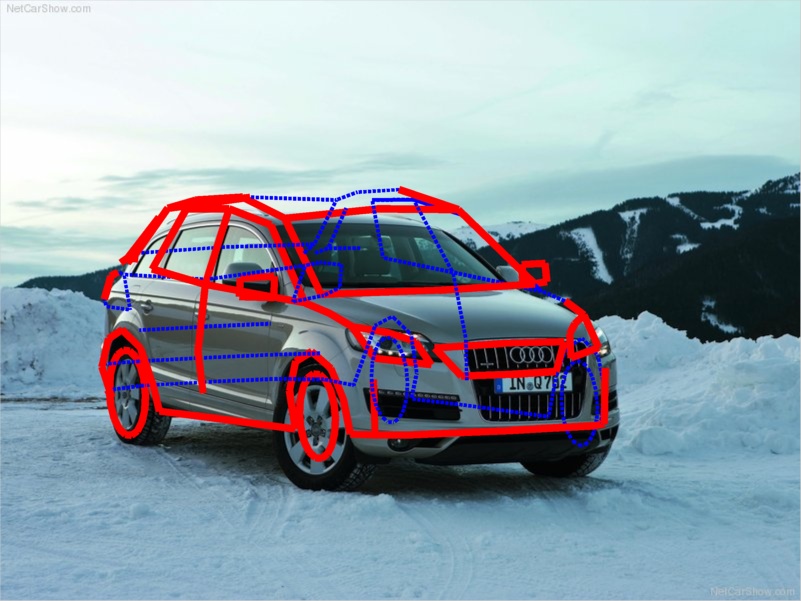}
        \end{subfigure}\\
        \begin{subfigure}[b]{0.25\textwidth}
                \includegraphics[width=\textwidth]{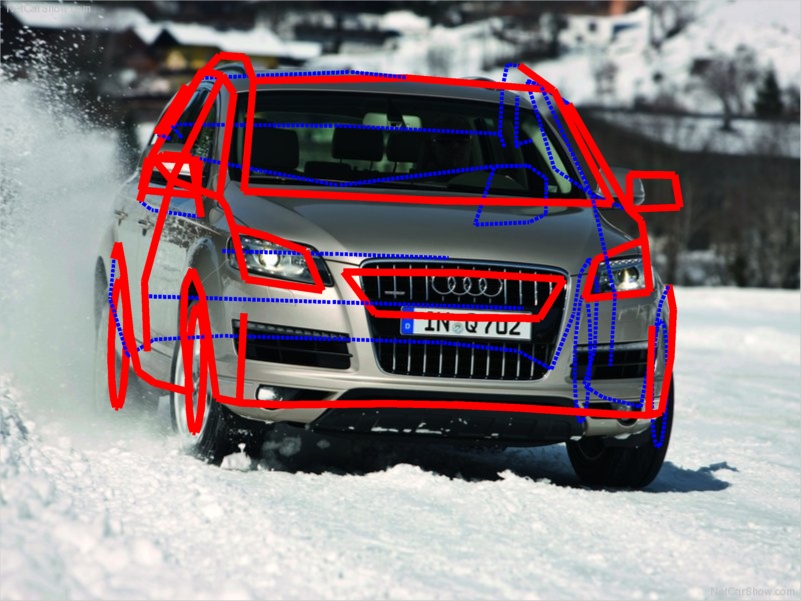}
        \end{subfigure}%
        ~ 
        \begin{subfigure}[b]{0.25\textwidth}
                \includegraphics[width=\textwidth]{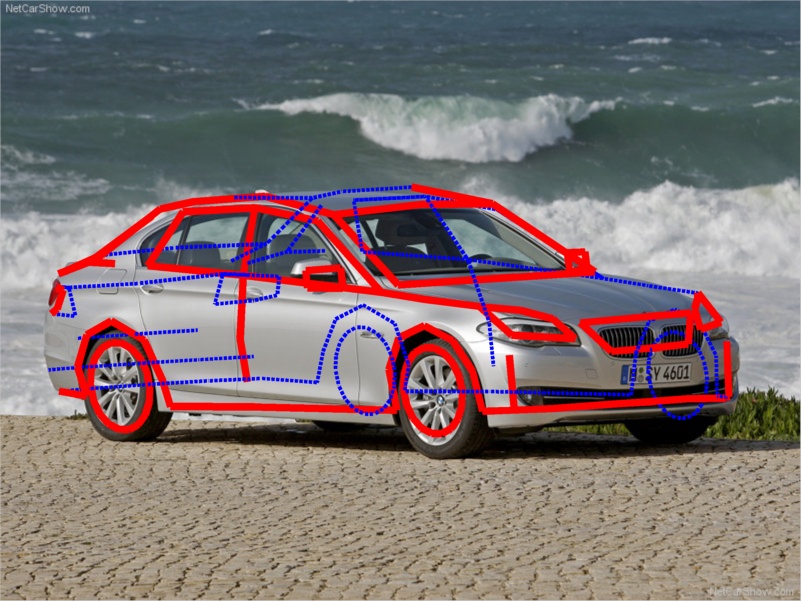}
        \end{subfigure}%
        ~
        \begin{subfigure}[b]{0.25\textwidth}
                \includegraphics[width=\textwidth]{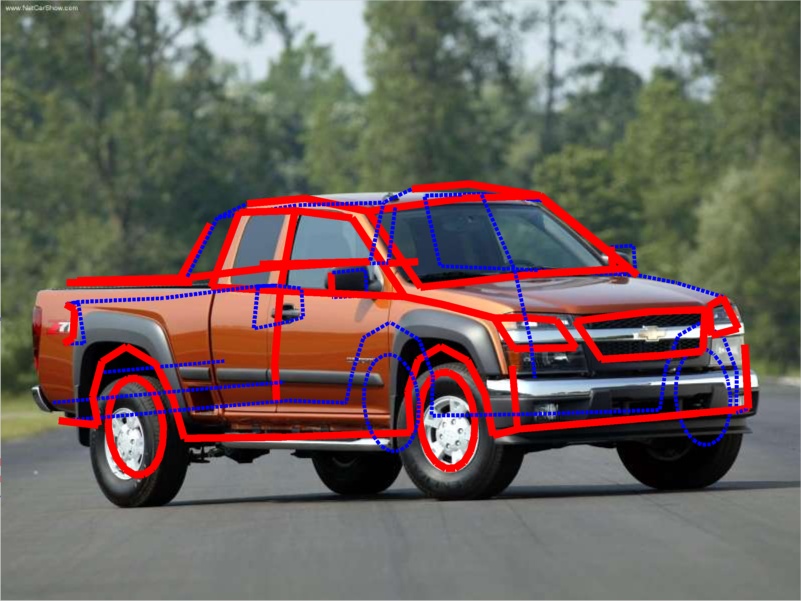}
        \end{subfigure}%
        ~ 
        \begin{subfigure}[b]{0.25\textwidth}
                \includegraphics[width=\textwidth]{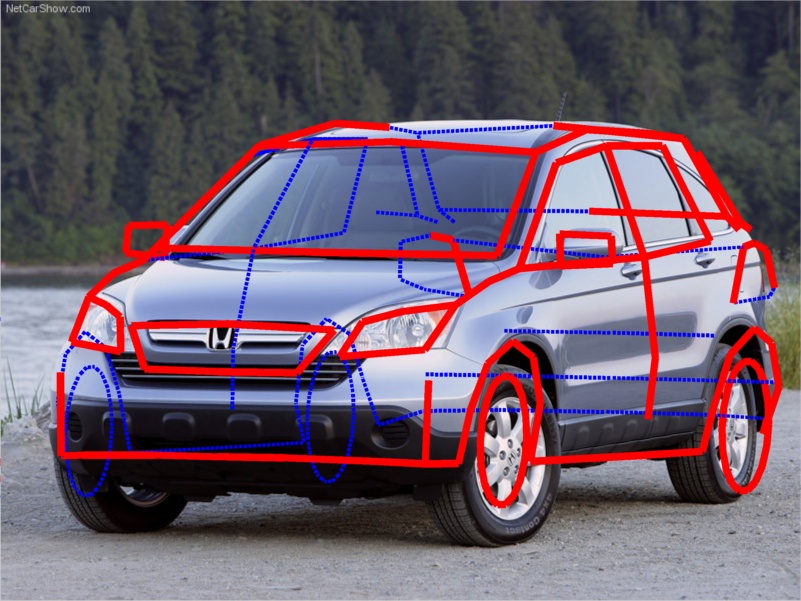}
        \end{subfigure}\\
        ~ 
        \begin{subfigure}[b]{\textwidth}
                \includegraphics[width=\textwidth]{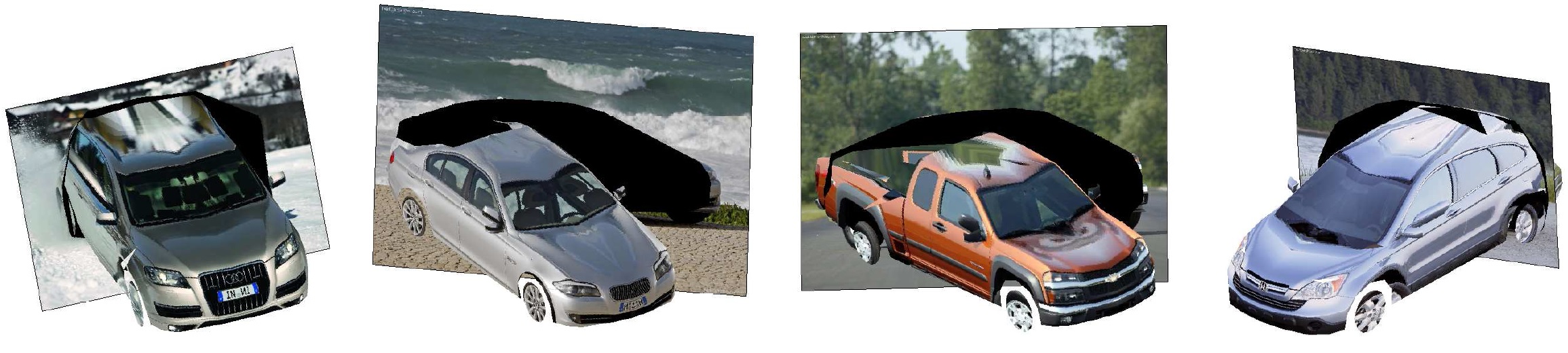}
        \end{subfigure}\\
        \caption{ On the first two rows, the 3D wire frame of the car model is projected on the image with estimated pose and shape. Red solid lines represent visible wire frames and blue dotted lines represent invisible wire frames. Our method robustly estimates the pose and shape for all car type and different view angles. On the last row, the textured 3D reconstructions of the cars on the fourth row are rendered at novel viewpoints. (We use symmetry to texture the invisible faces).}\label{fig:results}
\end{figure*}

In addition to the quantitative evaluations, we show qualitative results on the test images from FG3DCar in Figure \ref{fig:results}, where we project the 3D model wireframe with the estimated pose and shape on to the image. We also show the textured model rendered at novel views.




%% file: tex/conclusion.tex
We proposed a novel approach for estimating the pose and the shape of a 3D object from a single image. Our approach is based on a collection of automatically-selected and discriminatively-trained 2D parts with a 3D shape-space model to represent the geometric relation. In model inference, we simultaneously localized the parts, estimated the pose, and recovered the 3D shape by solving a convex program with ADMM. 

%% file: draft.bbl
\begin{thebibliography}{10}\itemsep=-1pt

\bibitem{aubry2014seeing}
M.~Aubry, D.~Maturana, A.~Efros, B.~Russell, and J.~Sivic.
\newblock Seeing 3d chairs: exemplar part-based 2d-3d alignment using a large
  dataset of cad models.
\newblock In {\em Proceedings of the IEEE Conference on Computer Vision and
  Pattern Recognition}, 2014.

\bibitem{boyd2010distributed}
S.~Boyd.
\newblock Distributed optimization and statistical learning via the alternating
  direction method of multipliers.
\newblock {\em Foundations and Trends in Machine Learning}, 3(1):1--122, 2010.

\bibitem{bregler2000recovering}
C.~Bregler, A.~Hertzmann, and H.~Biermann.
\newblock Recovering non-rigid 3d shape from image streams.
\newblock In {\em Proceedings of the IEEE Conference on Computer Vision and
  Pattern Recognition}, 2000.

\bibitem{cootes1995active}
T.~Cootes, C.~Taylor, D.~Cooper, and J.~Graham.
\newblock Active shape models -- their training and application.
\newblock {\em Computer Vision and Image Understanding}, 61(1):38--59, 1995.

\bibitem{cristinacce2006feature}
D.~Cristinacce and T.~Cootes.
\newblock Feature detection and tracking with constrained local models.
\newblock In {\em Proceedings of the British Machine Vision Conference}, 2006.

\bibitem{dalal2005histograms}
N.~Dalal and B.~Triggs.
\newblock Histograms of oriented gradients for human detection.
\newblock In {\em Computer Vision and Pattern Recognition, 2005. CVPR 2005.
  IEEE Computer Society Conference on}, volume~1, pages 886--893. IEEE, 2005.

\bibitem{felzenszwalb2010object}
P.~F. Felzenszwalb, R.~B. Girshick, D.~McAllester, and D.~Ramanan.
\newblock Object detection with discriminatively trained part-based models.
\newblock 32(9):1627--1645, 2010.

\bibitem{fidler20123d}
S.~Fidler, S.~Dickinson, and R.~Urtasun.
\newblock 3d object detection and viewpoint estimation with a deformable 3d
  cuboid model.
\newblock In {\em Advances in Neural Information Processing Systems}, 2012.

\bibitem{girshick2013training}
R.~Girshick and J.~Malik.
\newblock Training deformable part models with decorrelated features.
\newblock In {\em Proceedings of the International Conference on Computer
  Vision ({ICCV})}, 2013.

\bibitem{glasner2011viewpoint}
D.~Glasner, M.~Galun, S.~Alpert, R.~Basri, and G.~Shakhnarovich.
\newblock Viewpoint-aware object detection and pose estimation.
\newblock In {\em Proceedings of the International Conference on Computer
  Vision}, 2011.

\bibitem{grimsonbook}
W.~Grimson.
\newblock {\em Object recognition by computer: The role of geometric
  constraints}.
\newblock The MIT Press, Cambridge, MA, 1990.

\bibitem{gu2010discriminative}
C.~Gu and X.~Ren.
\newblock Discriminative mixture-of-templates for viewpoint classification.
\newblock In {\em Proceedings of the European Conference on Computer Vision},
  2010.

\bibitem{gu20063d}
L.~Gu and T.~Kanade.
\newblock {3D} alignment of face in a single image.
\newblock In {\em Proceedings of the IEEE Conference on Computer Vision and
  Pattern Recognition}, 2006.

\bibitem{hariharan2012discriminative}
B.~Hariharan, J.~Malik, and D.~Ramanan.
\newblock Discriminative decorrelation for clustering and classification.
\newblock In {\em Computer Vision--ECCV 2012}, pages 459--472. Springer, 2012.

\bibitem{hejrati2012analyzing}
M.~Hejrati and D.~Ramanan.
\newblock Analyzing 3d objects in cluttered images.
\newblock In {\em Advances in Neural Information Processing Systems}, 2012.

\bibitem{jiang2011linear}
H.~Jiang, S.~X. Yu, and D.~R. Martin.
\newblock Linear scale and rotation invariant matching.
\newblock {\em IEEE Transactions on Pattern Analysis and Machine Intelligence},
  33(7):1339--1355, 2011.

\bibitem{kokkinos2011rapid}
I.~Kokkinos.
\newblock Rapid deformable object detection using dual-tree branch-and-bound.
\newblock In {\em Advances in Neural Information Processing Systems}, pages
  2681--2689, 2011.

\bibitem{li2011optimal}
H.~Li, J.~Huang, S.~Zhang, and X.~Huang.
\newblock Optimal object matching via convexification and composition.
\newblock In {\em Proceedings of the International Conference on Computer
  Vision}, 2011.

\bibitem{liebelt2008viewpoint}
J.~Liebelt, C.~Schmid, and K.~Schertler.
\newblock Viewpoint-independent object class detection using 3d feature maps.
\newblock In {\em Proceedings of the IEEE Computer Vision and Pattern
  Recognition}, 2008.

\bibitem{lim2013parsing}
J.~J. Lim, H.~Pirsiavash, and A.~Torralba.
\newblock Parsing ikea objects: Fine pose estimation.
\newblock In {\em Proceedings of the International Conference on Computer
  Vision}, 2013.

\bibitem{Lin2014jointly}
Y.-L. Lin, V.~I. Morariu, W.~Hsu, and L.~S. Davis.
\newblock Jointly optimizing 3d model fitting and fine-grained classification.
\newblock In {\em Proceedings of the European Conference on Computer Vision},
  2014.

\bibitem{maciel2003global}
J.~Maciel and J.~P. Costeira.
\newblock A global solution to sparse correspondence problems.
\newblock {\em IEEE Transactions on Pattern Analysis and Machine Intelligence},
  25(2):187--199, 2003.

\bibitem{pepik20123d2pm}
B.~Pepik, P.~Gehler, M.~Stark, and B.~Schiele.
\newblock 3d2pm--3d deformable part models.
\newblock In {\em Proceedings of the European Conference on Computer Vision},
  2012.

\bibitem{pepik2012teaching}
B.~Pepik, M.~Stark, P.~Gehler, and B.~Schiele.
\newblock Teaching {3D} geometry to deformable part models.
\newblock In {\em Proceedings of the IEEE Conference on Computer Vision and
  Pattern Recognition}, 2012.

\bibitem{ramakrishna2012reconstructing}
V.~Ramakrishna, T.~Kanade, and Y.~Sheikh.
\newblock Reconstructing 3d human pose from 2d image landmarks.
\newblock In {\em Proceedings of the European conference on Computer Vision},
  2012.

\bibitem{savarese20073d}
S.~Savarese and F.-F. Li.
\newblock 3d generic object categorization, localization and pose estimation.
\newblock In {\em Proceedings of the IEEE International Conference on Computer
  Vision}, 2007.

\bibitem{schneiderman2000statistical}
H.~Schneiderman and T.~Kanade.
\newblock A statistical method for 3d object detection applied to faces and
  cars.
\newblock In {\em Proceedings of the IEEE Conference on Computer Vision and
  Pattern Recognition}, 2000.

\bibitem{Singh2012DiscPat}
S.~Singh, A.~Gupta, and A.~A. Efros.
\newblock Unsupervised discovery of mid-level discriminative patches.
\newblock In {\em European Conference on Computer Vision}, 2012.

\bibitem{sun2010depth}
M.~Sun, G.~Bradski, B.-X. Xu, and S.~Savarese.
\newblock Depth-encoded hough voting for joint object detection and shape
  recovery.
\newblock In {\em Proceedings of the European Conference on Computer Vision},
  2010.

\bibitem{sun2009multi}
M.~Sun, H.~Su, S.~Savarese, and L.~Fei-Fei.
\newblock A multi-view probabilistic model for 3d object classes.
\newblock In {\em Proceedings of the IEEE Conference on Computer Vision and
  Pattern Recognition}, 2009.

\bibitem{torralba2007sharing}
A.~Torralba, K.~P. Murphy, and W.~T. Freeman.
\newblock Sharing visual features for multiclass and multiview object
  detection.
\newblock {\em IEEE Transactions on Pattern Analysis and Machine Intelligence},
  29(5):854--869, 2007.

\bibitem{xiang_wacv14}
Y.~Xiang, R.~Mottaghi, and S.~Savarese.
\newblock Beyond pascal: A benchmark for 3d object detection in the wild.
\newblock In {\em IEEE Winter Conference on Applications of Computer Vision
  (WACV)}, 2014.

\bibitem{xiang2012estimating}
Y.~Xiang and S.~Savarese.
\newblock Estimating the aspect layout of object categories.
\newblock In {\em Proceedings of the IEEE International Conference on Computer
  Vision and Pattern Recognition}, 2012.

\bibitem{yan20073d}
P.~Yan, S.~M. Khan, and M.~Shah.
\newblock 3d model based object class detection in an arbitrary view.
\newblock In {\em Proceedings of the International Conference on Computer
  Vision}, 2007.

\bibitem{zhou2014sptio}
F.~Zhou and F.~De~la Torre.
\newblock Spatio-temporal matching for human detection in video.
\newblock In {\em Proceedings of the European Conference on Computer Vision},
  2014.

\bibitem{zhou20143d}
X.~Zhou, S.~Leonardos, X.~Hu, and K.~Daniilidis.
\newblock 3d shape reconstruction from 2d landmarks: A convex formulation.
\newblock {\em Arxiv preprint arXiv: 1411.2942}, 2014.

\bibitem{zia2013detailed}
M.~Z. Zia, M.~Stark, B.~Schiele, and K.~Schindler.
\newblock Detailed 3d representations for object recognition and modeling.
\newblock {\em IEEE Transactions on Pattern Analysis and Machine Intelligence},
  35(11):2608--2623, 2013.

\end{thebibliography}
